\documentclass{article}
\usepackage{spconf,amsmath,graphicx}
\usepackage{color}
\usepackage{multirow}
\usepackage{tikz}
\usepackage{pgfplots}
\pgfplotsset{compat=1.17}
\usepackage{enumitem}

\def\F{\mathcal{F}}
\def\G{\mathcal{G}}
\def\bx{\textbf{x}}

\title{COMPACT AND ADAPTIVE MULTIPLANE IMAGES FOR VIEW SYNTHESIS}
\name{Julia Navarro and Neus Sabater}
\address{InterDigital}

\begin{document}
%
\maketitle
\begin{abstract} 
Recently, learning methods have been designed to create Multiplane Images (MPIs) for view synthesis. While MPIs are extremely powerful and facilitate high quality renderings, a great amount of memory is required, making them impractical for many applications. In this paper, we propose a learning method that optimizes the available memory to render compact and adaptive MPIs. Our MPIs avoid redundant information and take into account the scene geometry to determine the depth sampling.

\end{abstract}
%

\section{Introduction} \label{sec:intro}

The emergence of Light Fields and volumetric video has created a new opportunity to provide compelling immersive user experiences. In particular, the sense of depth and parallax in real scenes provides a high level of realism. 
It turns out that MPIs, a stack of semitransparent images, are a handy volumetric representation to synthetize new views. Indeed, MPIs excel at recovering challenging scenes with transparencies or reflective surfaces and complicated occlusions with thin objects. Furthermore, once the MPI is computed, rendering several virtual views can be done very efficiently with angular consistency and without flickering artifacts.

Lately, many works have focused on the learning of MPIs. In \cite{zhou2018stereo} the MPI from two views with narrow baseline to extrapolate novel images is computed. \cite{srinivasan2019pushing} proposes a two-stage method, in which the MPI is filtered in a second step recomputing color and alpha values of occluded voxels through optical flow estimation. \cite{mildenhall2019local} averages renderings obtained from MPIs of nearby viewpoints. \cite{flynn2019deepview} introduced a learned gradient descent scheme to iteratively refine the MPI. Network parameters are not shared per iteration and their approach is computationally expensive.  This method is applied to multi-view videos in \cite{broxton2020immersive}. A lighter model is proposed in \cite{volker2020learning} taking iterative updates on the alpha channel, with shared weights per iteration. The MPI colors are extracted from the input Plane Sweep Volumes (PSVs) using visibility cues provided by the estimated alpha.

However, MPIs are bulky and their memory footprint make them prohibitive for many applications. In particular, the bottleneck of deep learning approaches to generate MPIs is the amount of required RAM for both training and inference. 
In this paper we present a novel view synthesis learning approach that computes compact and adaptive MPIs. This is, our MPIs 
(i)	are as compact as possible and avoid redundant information, so they are easily compressible; and 
(ii) have a depth sampling adapted to the given scene, meaning that depth planes are placed at the scene objects depths.
With our strategy we leverage the MPI representation while we break through the constraint of memory. Indeed, for a given available data volume, while other methods compute the fill-in values of the fixed MPI, our solution computes both the container and the content that better fits the scene. Besides, thanks to the compactness of our MPIs, we guarantee a great compression capability without losing image quality which is of foremost importance for volumetric data  transmittance.%

\section{Proposed method} \label{sec:proposed}
An MPI is a set of $D$ fronto-parallel planes placed at different depths $\mathbf{d} = d_1, \dots, d_D$ (in back-to-front order), with respect to a reference camera. Each plane $d$ consists of an RGB$\alpha$ image that encodes  color and a transparency/opacity value $\alpha_d(\bx)$, for each pixel $\bx = (x,y)$ in the image domain $\Omega$. Given an MPI, novel views can be rendered by warping each image plane and alpha compositing \cite{zhou2018stereo}. In this paper, we aim at computing an MPI given $K\geq2$ input images $I_1, \dots, I_K$, with its associated camera parameters. Since the operations to render novel views from an MPI are differentiable, a learning-based solution can be supervised with the final synthesized views and no MPI ground truth is required.
\vspace{-0.2cm}
\subsection{Learning compact MPIs}
Inspired by \cite{flynn2019deepview,volker2020learning}, we compute the MPI iteratively
\begin{equation}\label{eq:itermpi}
M_{n+1} = \mathcal{S}\left(M_n + \F(H_n, M_n)\right),
\end{equation}
where $\mathcal{S}$ is a sigmoid function and $\F$ is a CNN that outputs the RGB$\alpha$ residual from the input features $H_n$ and the previous iteration MPI $M_n$. In contrast to \cite{flynn2019deepview} and similar to \cite{volker2020learning}, we share the weights of this network across iterations. 
In particular, $H_n = (\bar{v}_n, \mu_n, \sigma_n^2, F_n)$ is a concatenation of visual cues \cite{volker2020learning} and deep features that enable our model to be applied to any number of input views arranged in any order.
$
F_n = \max_k \left\{ \mathcal{G}(P_k, M_n) \right\}
$
are input deep features at iteration $n$, where $\G$ is a CNN with shared weights between all views and iterations and $P_k$ is the PSV of view $k$. Also, $\bar{v}_n$ is the total visibility, that computes how many cameras see each voxel;  $\mu_n$ is the mean visible color, an across-views average of the PSVs of the input images weighted by the view visibility; and  $\sigma_n^2$ is the visible color variance, that measures how the mean colors differ from the PSVs at visible voxels.

Regarding the initial MPI $M_0$, the RGB$\alpha$ planes are selected such that the color channels are equal to the focal stack, while the alpha component consists of a transparent volume with an opaque background plane. Note that we could only input $P_k$ to $\G$, but the concatenation with $M_n$ is beneficial to identify relations between PSVs and the MPI.

\begin{figure*}[t!]\centering
\includegraphics[trim=0 418 80 0,clip=true,width=\linewidth]{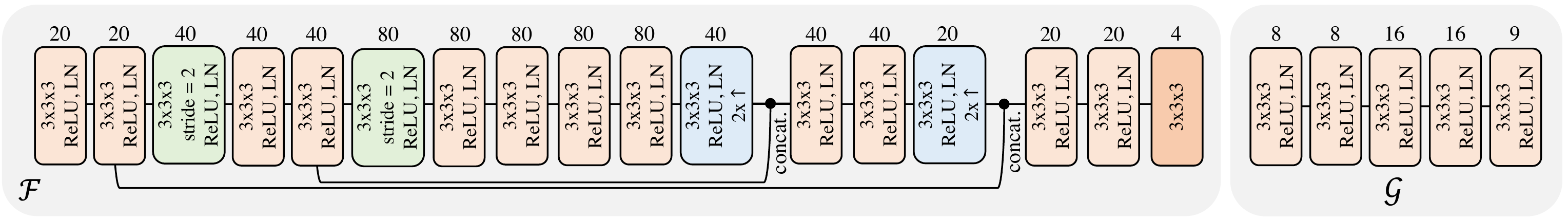}
\caption{Networks $\F$ and $\G$. We use 3D CNNs with a ReLU activation and layer normalization (LN) \cite{ba2016layer}, except the last layer in $\F$. We use kernel sizes of $3\times3\times3$ and zero-padding to keep spatial and depth dimensions. Layers in green include a stride of 2 pixels in the three dimensions. Layers in blue apply $2\times$ bilinear upsampling in spatial and depth dimensions.}
\label{fig:networks}
\end{figure*}

\textbf{Networks architectures.} 
We use 3D convolutions in $\F$ and $\G$. This allows to compute MPIs with variable spatial and depth dimensions. Fig. \ref{fig:networks} details the two architectures. Similar to \cite{volker2020learning}, $\F$ is a 3D UNet network \cite{ronneberger2015u} with a last 3D CNN layer.

\textbf{Training loss.}
To supervise the rendering quality, we consider $V\geq1$ views $I_1, \dots, I_V$, which are different from the $K$ input images. {\color{black} The main term that guides the training is the feature distance used in \cite{flynn2019deepview} between ground truth views $I_v$ and views $\hat{I}_v$ rendered from the estimated MPI. We denote this perceptual loss $\mathcal{L}_p$ and could be the only term.} However, with no other constraint, it tends to provide MPIs with many voxels with positive alpha values. This results in unnecessary information that is repeated in different planes. Generally, most of this data is not reflected in the rendered views since it ends up covered by foreground voxels after the over-compositing operation. Then, we seek to provide an optimal representation that does not encode redundant information while it produces high quality rendered views. We propose to promote a compact MPI by means of a sparsity loss term that limits the number of voxels with non-zero alphas. In particular, we minimize 
\begin{equation}\label{eq:sparseMPI}
\hat{A} = \sum_{\bx\in\Omega} \max\{A(\bx)-\tau(\bx), 0\},
\end{equation}
where $A(\bx) = \sum_{d=1}^D \alpha_d(\bx)$ is the image of accumulated alphas, an approximation of how many planes are activated for each pixel, and  $\tau(\bx)$ is the number of allowed activated planes per pixel. Note that when $A(\bx)<\tau(\bx)$ the maximum is zero, meaning that we allow up to $\tau(\bx)$ planes with zero cost in the loss. Now, $\tau$ is automatically computed from the total visibility $\bar{v}_n$. Indeed, for each $\bx$, we inspect the values of the vector $\bar{v}_n(\bx)$ along the depth dimension. If there is a voxel that is visible only by a subset of cameras (i.e., $1<\bar{v}_n(\bx)<K$), it means that it is a semi-occluded region, and the pixel should be encoded with more active depth planes in the MPI. In that case $\tau(\bx)=6$ and $\tau(\bx)=3$ otherwise. 

With the term in Eq. \eqref{eq:sparseMPI} in the loss, the network produces compact MPIs. However, there are cases in which $A$ does not reach a minimum value of $1$ for some pixels. This mostly happens at disocclusion pixels when the MPI planes are warped to other views different than the reference one. To prevent this, we also enforce $A$ to have a minimum value of one, for the reference camera and to the other input views after warping the alpha planes. If we denote $A_\text{min}$ as the minimum of $A$ over all reference and $K$ input views, the sparsity loss term is
\begin{equation}
 \mathcal{L}_s = \frac{1}{|\Omega|} \hat{A}
 +|\min\{A_{\text{min}} -1, 0\}|.
\end{equation}

Then, we train our method with the loss $\mathcal{L}= \mathcal{L}_p + \lambda\,\mathcal{L}_s$, a combination of synthesis quality and MPI compactness, weighted by $\lambda$  (experimentally set to $0.1$).

\textbf{Training details.} 
Our model is implemented with TensorFlow and trained end-to-end with ADAM \cite{kingma2015adam}  (default parameters), Xavier initialization \cite{glorot2010understanding} and updated with a batch size of 1, on $92\times92$ patches with $D=60$. We increase the number of MPI refinement steps with the training iterations (2 for 0-50k, 3 for 50k-100k and 4 for 100k-215k).
The training lasts 8 days on a Titan RTX GPU.

\subsection{Scene-adapted depth sampling}
In the literature \cite{zhou2018stereo,srinivasan2019pushing,mildenhall2019local,flynn2019deepview,volker2020learning}, the MPI planes are equally spaced according to inverse depth in a fixed depth range, which is globally selected to be coherent with the whole dataset. However, both the depth range and the regular sampling may not be optimal for every scene, so the resulting MPIs may have empty depth slices. These planes occupy memory while not being meaningful for the scene that is represented. Instead, we propose to adapt the depth sampling to the given scene by redistributing these empty slices and placing them at more relevant depths. Consequently, scene objects will be located at more accurate depths, leading to synthesized views with higher quality. Given the regular list of depths $\mathbf{d}$, an adaptive MPI is computed as follows:
\begin{enumerate}[label=\arabic*)]
    \item {\it Localize and discard irrelevant depths}. We compute $M_1$ with depths $\mathbf{d}$ (one iteration of Eq. \eqref{eq:itermpi}), since it contains already the scene geometry, although refined in further iterations. Then, we remove the depths in $\mathbf{d}$ which corresponding $\alpha$ channel does not reach a minimum threshold for any pixel (set in practice to $0.3$).
    
    \item {\it Assign weights to remaining depth intervals}. We assign to each depth plane a weight based on the spatial average of its corresponding $\alpha$. The weight of each depth interval is the average of the weights of its endpoints.
    
    \item {\it Distribute new depths}. We place as many depths as removed in 1) at intervals with higher weights. Depth within an interval is regularly sampled on the inverse.
    
    \item {\it Compute the scene-adapted MPI}. We recompute the PSVs with the new depth sampling and apply our iterative method from the beginning.
\end{enumerate}

Notice that this process is only applied at inference. Our experiments with this module at training show slower optimization with no significant improvements. Also, the identification of empty planes from the alpha channel works since our network outputs a compact MPI representation.

\section{Experiments} \label{sec:experiments}

In this section, we assess the performance of our approach. Quantitative evaluation is in terms of SSIM \cite{wang2004image} (higher is better) and LPIPS \cite{zhang2018unreasonable} (lower is better). 
We refer to our supplemental material for results on multi-view videos, in which we applied our method to each frame.

\textbf{Datasets.} 
We use the InterDigital (ID) \cite{sabater2017dataset} and the Spaces \cite{flynn2019deepview} datasets. We consider the same augmented ID dataset used in \cite{volker2020learning} made of 21 videos to train and 6 to test, captured with a $4\times4$ camera rig.  During training, we select $K=4$ random views of a random $3\times3$ subrig, while the remaining $V=5$ are used for supervision. 
We consider a higher spatial resolution than in \cite{volker2020learning}, with $1024\times544$ pixels. The Spaces dataset consists of 90 training and 10 test scenes, captured with a 16-camera rig. We consider the {\it small baseline} configuration which resolution is $800\times480$, with fixed $K=4$ and $V=3$. When testing different configurations of our system, we use the ID data with resolution $512\times272$ and $D=32$ instead of 60 for both train and test. In all cases, the MPI reference camera is computed from an average of the positions and orientations of input cameras \cite{markley2007averaging}.

\begin{table}[t!]\centering
{\renewcommand{\arraystretch}{1.3}
\resizebox{\columnwidth}{!}{
\begin{tabular}{|c|c|c|c|c|}\cline{2-5}
\multicolumn{1}{c|}{}& $H_n=F_n$ & $\G(P_k)$ & Reg. sampl. & Proposed \\\hline
SSIM $\uparrow$ & 0.9289 & 0.9414 & 0.9391 & 0.9441 \\ \hline
LPIPS $\downarrow$ & 0.0465 & 0.0346 & 0.0392 & 0.0334\\ \hline
\end{tabular}
}}
\caption{Analysis of the inputs to networks $\F$ and $\G$, our approach with regular depth sampling, and our complete method with the proposed adaptive one. Metrics are averaged over the ID test set.}
\label{tab:ablation}
\end{table}

\textbf{Networks inputs.}
Table \ref{tab:ablation} reports the results obtained when only using as input to $\F$ the deep features $F_n$. Results significantly improve when $F_n$ is reinforced with the visual cues $\mu_n$, $\sigma^2_n$ and $\bar{v}_n$. Table \ref{tab:ablation} also compares the case of only feeding $\G$ with the PSV $P_k$. In this case, $\G$ requires less training parameters and it is no longer part of the iterative loop, leading to a reduction of the computational cost, but the inclusion of the MPI as input increases the quality of synthesized views.

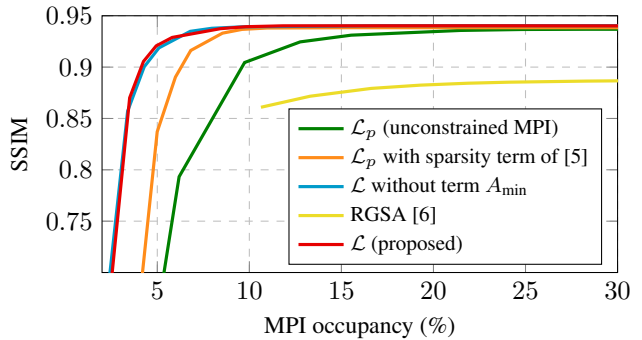
\begin{figure}[t!]\centering
\begin{tikzpicture}
\begin{axis}[
    xlabel={\small MPI occupancy (\%)},
    ylabel={\small SSIM}, 
    xmin=2, xmax=30, 
    ymin=0.7, ymax=0.95,
    xtick={5,10,15,20,25,30,35},
    ytick={0.75,0.80,0.85,0.90,0.95}, 
    legend pos=south east, 
    legend cell align={left},
    ymajorgrids=true,
    xmajorgrids=true,
    grid style=dashed,
    y post scale=0.6,
    every axis plot/.append style={very thick}
]

\addplot[
    color=green!50!black,
    mark size=0.8pt
    ]
    coordinates {
	(100.0,0.9367862)(38.94,0.9367862)(36.64,0.936806)(28.38,0.9368409)(24.81,0.9366289)(21.47,0.93570981)(15.53,0.9310923)(12.76,0.9245518)(9.743,0.90447087)(6.186,0.793270)(1.390,0.2483639)(0.200,0.062222) 

	};
    \addlegendentry{\footnotesize $\mathcal{L}_p$ (unconstrained MPI)}
	
\addplot[
    color=orange!90!white,
    mark size=0.8pt
    ]
    coordinates {
	(100,0.93835)(19.354,0.938353)(15.602,0.938377)(10.96,0.938064)(9.606,0.9367350)(8.543,0.933064)(6.818, 0.9162005)(5.986,0.8902839)(4.996,0.8370533)(3.753,0.6237237)(2.137,0.227655)(1.461,0.067988)

	};
	\addlegendentry{\footnotesize $\mathcal{L}_p$ with sparsity term of \cite{broxton2020immersive}}

\addplot[
    color=cyan!80!black,
    mark size=0.8pt
    ]
    coordinates {
	(100,0.940344494)(21.74,0.940344)(17.11,0.940334)(9.59,0.9391973)(7.96,0.937800)(6.79,0.934818)(5.09,0.918728)(4.29,0.900299)(3.39,0.857181)(2.31,0.680215)(0.78,0.247101)(0.21,0.0733241)

    };
    \addlegendentry{\footnotesize $\mathcal{L}$ without term $A_\text{min}$}
    
\addplot[
    color=yellow!90!black,
    mark size=0.8pt
    ]
    coordinates {
     (100.0,0.886847)(35.75,0.886847)(35.14,0.886856)(33.38,0.88687)(30.71,0.886726)(28.30,0.886375)(24.10,0.885323)(21.87,0.884313)(19.29,0.882442)(16.62,0.879258)(13.30,0.871645)(10.63,0.860895)
    };
    \addlegendentry{\footnotesize RGSA \cite{volker2020learning}}
    
\addplot[
    color=red!90!black,
    mark size=0.8pt
    ]
    coordinates {
	(100.0,0.940157)(30.53,0.9401572)(22.31,0.940269)(11.82,0.940186)(9.817,0.939344)(8.43,0.9374247)(5.81,0.9288969)(4.94,0.9206481)(4.25, 0.905485)(3.50,0.8701525)(2.34,0.6621339)(1.20, 0.353448) 

	};
    \addlegendentry{\footnotesize $\mathcal{L}$ (proposed)}
\end{axis}
\end{tikzpicture}
\caption{MPI occupancy against SSIM when thresholding the alpha channel under different values in $[0,0.95]$. A curve standing to the top left of another curve is preferred.}
\label{fig:graphocc}
\end{figure}

\textbf{Compactness evaluation.} 
Transparent voxels in the MPI ideally have a zero $\alpha$, but in practice, it is not the case. Setting to zero the $\alpha$ channel for values smaller than a threshold is a required step to reduce the memory footprint and only encode the voxels that are essential to obtain high quality renderings.
Varying the threshold leads to different MPI occupancies and different reconstruction qualities. Fig. \ref{fig:graphocc} illustrates this trade-off for our approach with our loss $\mathcal{L}$, other loss functions and for the recurrent geometry segmentation approach~\cite{volker2020learning} (RGSA). Our loss achieves the best compromise between sparsity and SSIM. 
In fact, our SSIM is stabilized with less than $10\%$ of voxels with non-zero alphas, while for the unconstrained version, it is achieved only with an occupancy close to $30\%$. In \cite{broxton2020immersive} the pixel-wise sum of the ratio between the $L^1$ and $L^2$ norms of the vector gathering the alpha values along the depth dimension is added to the synthesis error to achieve sparsity, but there is no mention about its weight. We have seen that the best results are achieved with a weight of $10^{-4}$, but with this loss the performance drops sooner than with ours. That is, with lower occupancy rates we sacrifice less in terms of quality.

\begin{figure}[t!]\centering
{\renewcommand{\arraystretch}{0.8}
\begin{tabular}{@{}c@{\hskip0.2em}c@{\hskip0.2em}c@{\hskip0.2em}c@{}}
\includegraphics[width=0.24\linewidth,trim={170 330 754 114},clip=true]{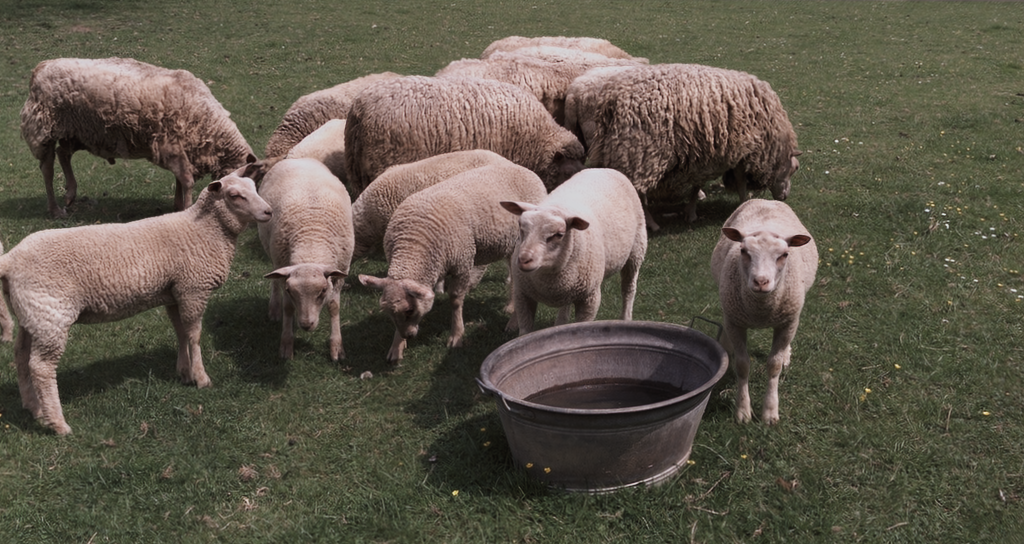} &
\includegraphics[width=0.24\linewidth,trim={170 330 754 114},clip=true]{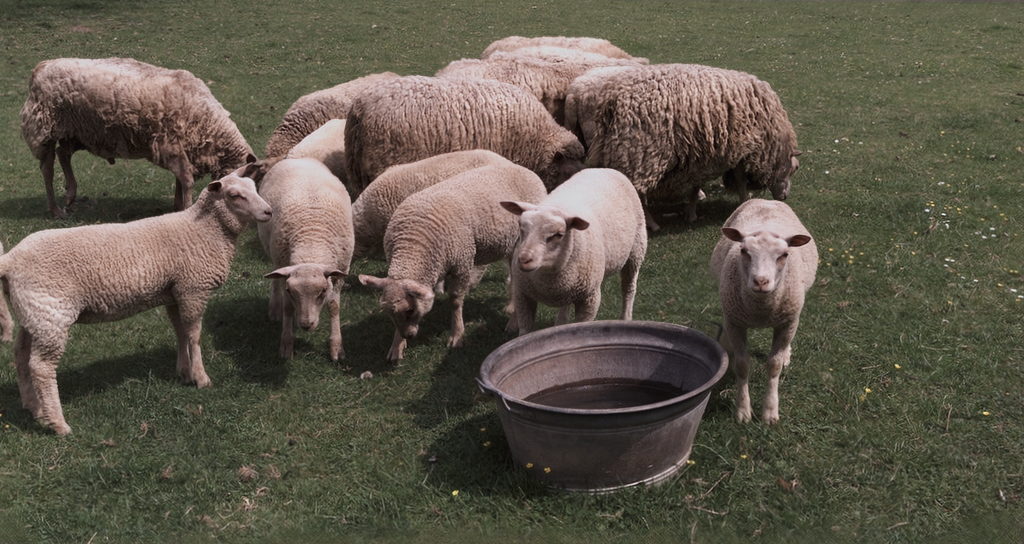} &
\includegraphics[width=0.24\linewidth,trim={600 370 324 74},clip=true]{figures/SheepsWithWater/SheepsWithWater_Ours-D_gt5_crops4.png} &
\includegraphics[width=0.24\linewidth,trim={600 370 324 74},clip=true]{figures/SheepsWithWater/SheepsWithWater_ours-a_gt5_crops4.png} 
\\
\includegraphics[width=0.24\linewidth,trim={770 244 154 200},clip=true]{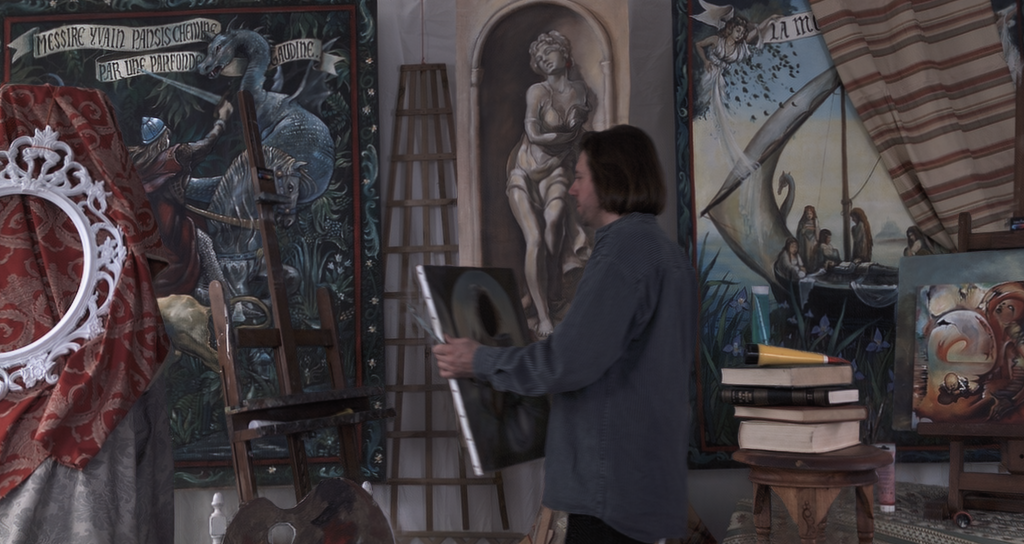} &
\includegraphics[width=0.24\linewidth,trim={770 244 154 200},clip=true]{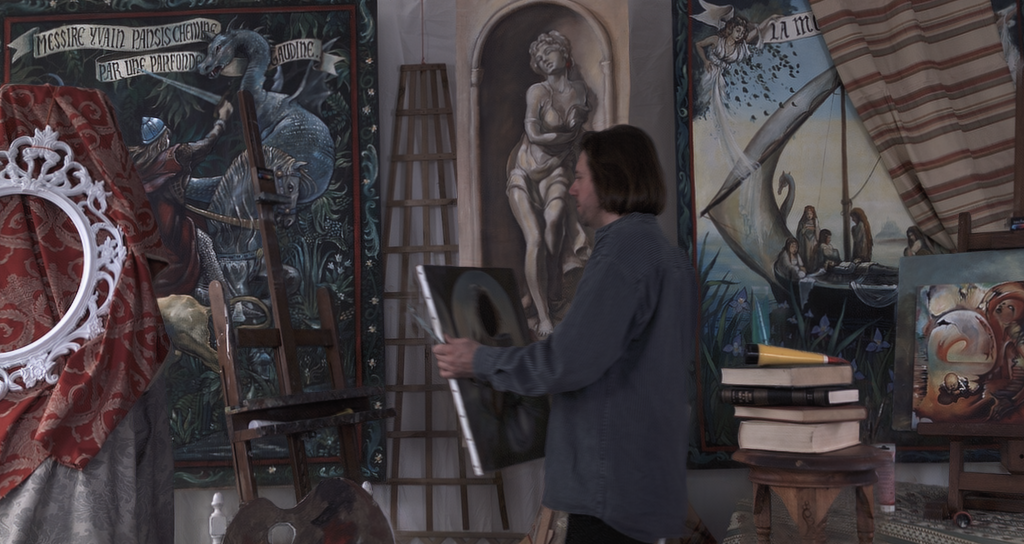} &
\includegraphics[width=0.24\linewidth,trim={720 444 204 0},clip=true]{figures/Painter/Painter_Ours-D_gt5_crops4.png} &
\includegraphics[width=0.24\linewidth,trim={720 444 204 0},clip=true]{figures/Painter/Painter_ours-a_gt5_crops4.png} 
\\
\small Regular & \small Proposed & \small Regular & \small Proposed
\end{tabular}
}
\caption{\color{black} With our proposed depth sampling we obtain sharper synthesized views compared to the use of a simple regular partition of the disparity. We refer to our supplementary material to better notice the differences.}
\label{fig:sampling}
\end{figure}

\textbf{Adaptive depth sampling.} 
With our strategy we obtain a finer depth partition of depth intervals containing scene objects. This leads to superior renderings without the need of increasing the number of MPI planes. Table \ref{tab:ablation} reports quantitative improvements of our proposed adaptive over regular sampling, while in Fig.~\ref{fig:sampling} we visually compare two examples of the ID test set. All crops are sharper, and textures are better preserved with the adaptive sampling.

\begin{table}[t!]\centering
{\renewcommand{\arraystretch}{1.3}
\resizebox{\columnwidth}{!}{
\begin{tabular}{|c|c|c|c|c|c|}\cline{3-6}
\multicolumn{2}{c|}{}& Soft3D	&  DeepView	& RGSA &  Proposed \\ \hline
\multirow{2}{*}{ID}& \small SSIM $\uparrow$& - & - & 0.8799 &  0.9150 \\\cline{2-6}
& \small LPIPS $\downarrow$ & -& - & 0.1748 & 0.0651 \\ \hline
\multirow{2}{*}{Spaces}& \small SSIM $\uparrow$ & 0.9261 & 0.9555 & 0.9084 & 0.9483\\ \cline{2-6}
& \small LPIPS $\downarrow$ & 0.0846 & 0.0343 & 0.1248 & 0.0453\\ \hline
\end{tabular}
}}
\caption{\color{black} Comparison with the state of the art. Metrics are averaged over the evaluation views of each test set. } 
\label{tab:sota} 
\end{table}

\begin{figure}[t!]\centering
{\renewcommand{\arraystretch}{0.8}
\begin{tabular}{@{}c@{\hskip0.4em}c@{}}
\includegraphics[width=0.49\linewidth]{figures/Painter/Painter_ours-a_gt5_crops4.png} &
\includegraphics[width=0.49\linewidth]{figures/TristanWannaPlayTennis/TristanWannaPlayTennis_ours-a_gt5_crops4}
\\
\end{tabular}
\begin{tabular}{@{}c@{\hskip0.2em}c@{\hskip0.2em}c@{\hskip0.4em}c@{\hskip0.2em}c@{\hskip0.2em}c@{}}
\includegraphics[width=0.158\linewidth,trim={350 250 574 194},clip=true]{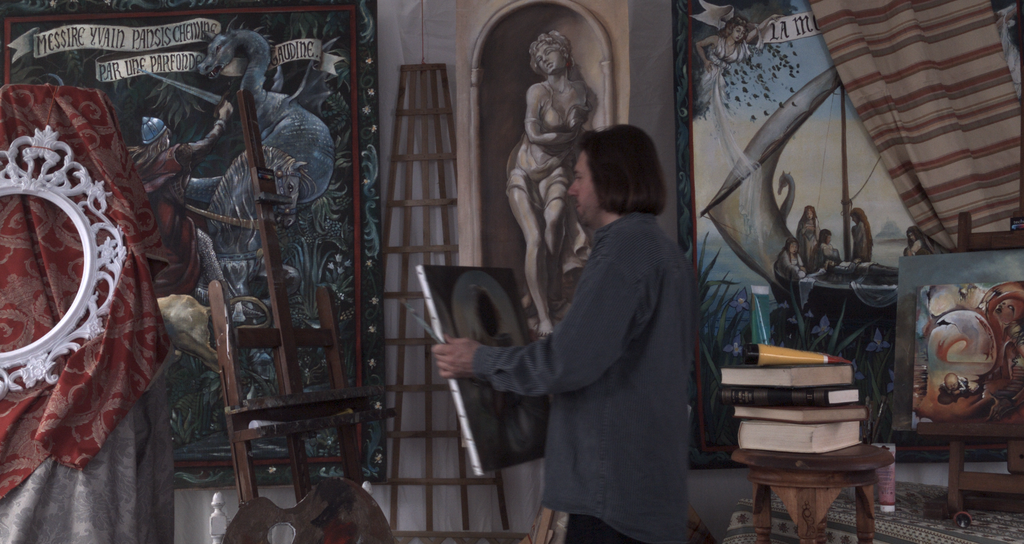} &
\includegraphics[width=0.158\linewidth,trim={350 250 574 194},clip=true]{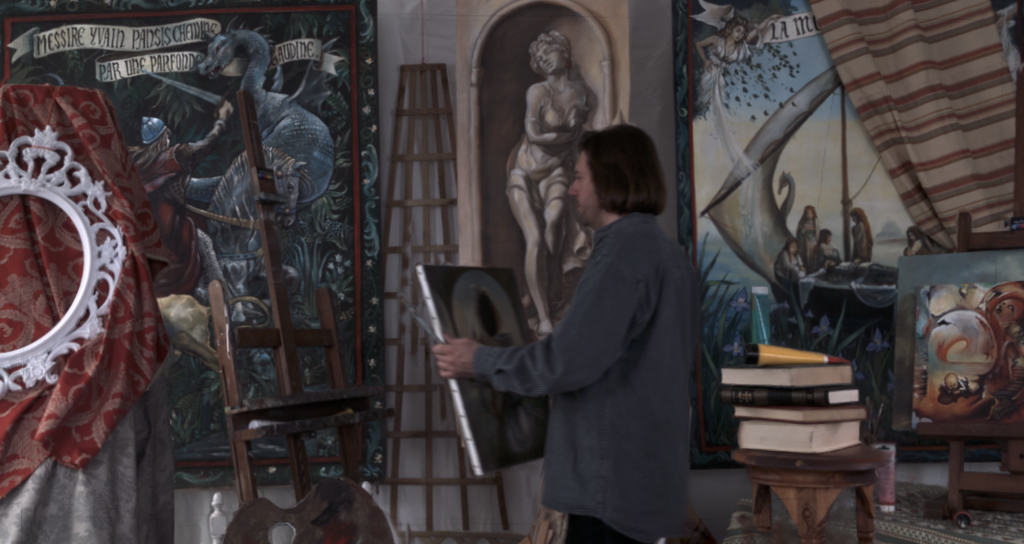} &
\includegraphics[width=0.158\linewidth,trim={350 250 574 194},clip=true]{figures/Painter/Painter_ours-a_gt5_crops4.png} &
\includegraphics[width=0.158\linewidth,trim={650 20 274 424},clip=true]{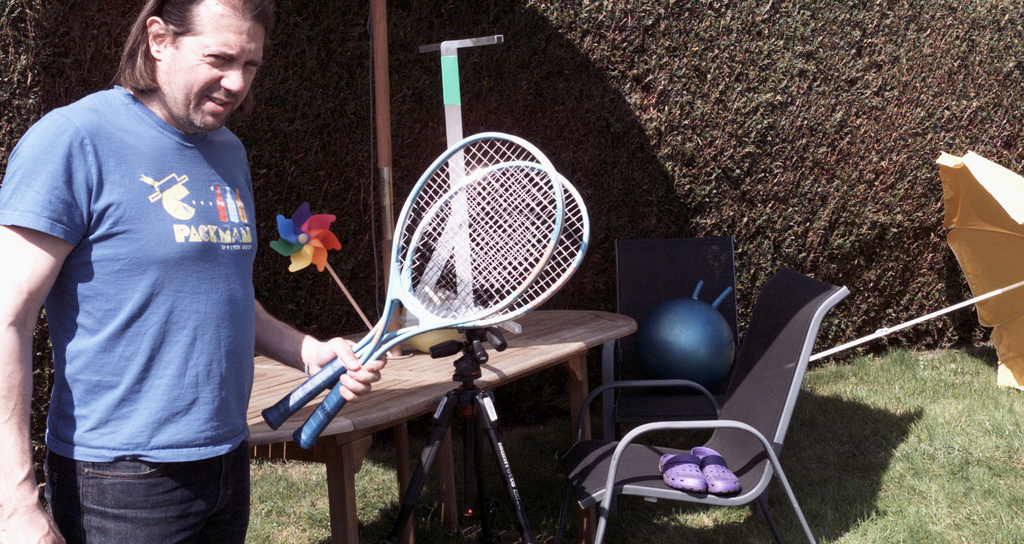} &
\includegraphics[width=0.158\linewidth,trim={650 20 274 424},clip=true]{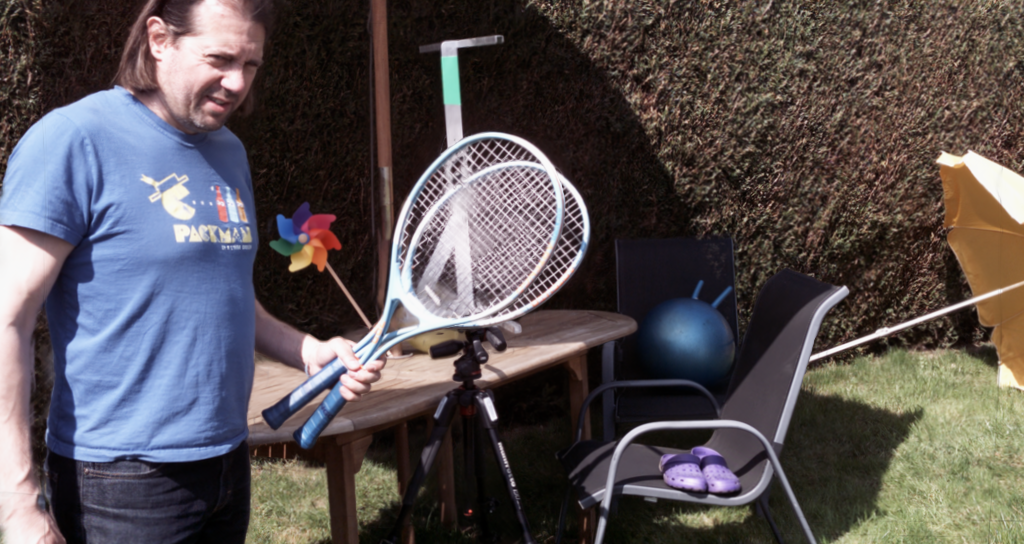} &
\includegraphics[width=0.158\linewidth,trim={650 20 274 424},clip=true]{figures/TristanWannaPlayTennis/TristanWannaPlayTennis_Ours-A_gt5_crops4.png} 
\\
\includegraphics[width=0.158\linewidth,trim={200 200 724 244},clip=true]{figures/Painter/Painter_gt5.png} &
\includegraphics[width=0.158\linewidth,trim={200 200 724 244},clip=true]{figures/Painter/Painter_volker_gt5_crops2.png} &
\includegraphics[width=0.158\linewidth,trim={200 200 724 244},clip=true]{figures/Painter/Painter_Ours-D_gt5_crops4.png} &
\includegraphics[width=0.158\linewidth,trim={420 320 504 124},clip=true]{figures/TristanWannaPlayTennis/TristanWannaPlayTennis_gt5.png} &
\includegraphics[width=0.158\linewidth,trim={420 320 504 124},clip=true]{figures/TristanWannaPlayTennis/TristanWannaPlayTennis_volker_gt5_crops2.png} &
\includegraphics[width=0.158\linewidth,trim={420 320 504 124},clip=true]{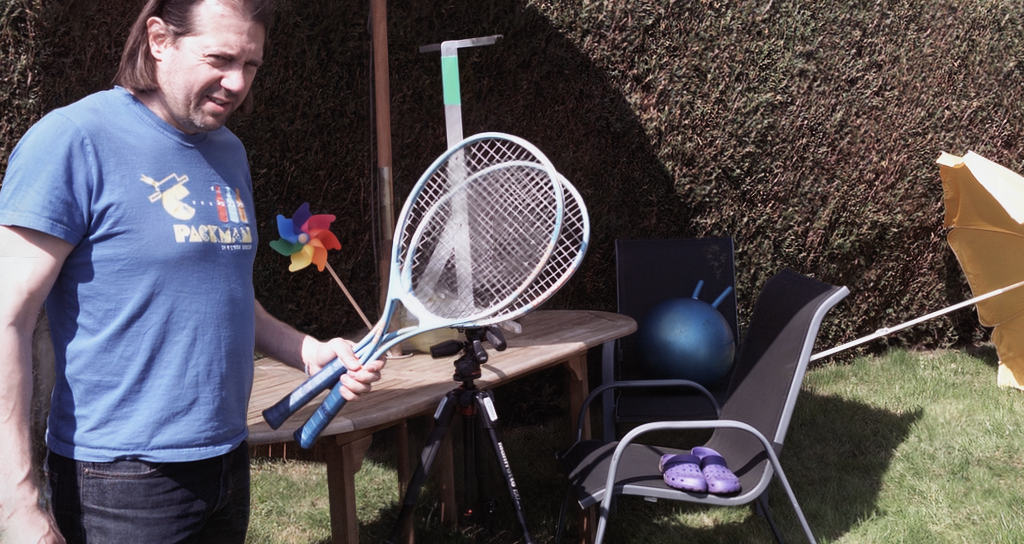} 
\\
\small GT & \small RGSA \cite{volker2020learning} & \small Proposed &
\small GT &\small  RGSA \cite{volker2020learning} & \small Proposed
\end{tabular}
}
\caption{\color{black} Synthesized central views of two examples of the ID test set. Result provided by our approach (top) and crops comparing with the ground truth (GT) and RGSA \cite{volker2020learning} (bottom).}
\label{fig:comparison_id}
\end{figure}

\begin{figure}[t!]\centering
{\renewcommand{\arraystretch}{0.8}
\begin{tabular}{@{}c@{\hskip0.4em}c@{\hskip0.4em}c@{}}
\includegraphics[width=0.3\linewidth,trim={300 30 300 234},clip=true]{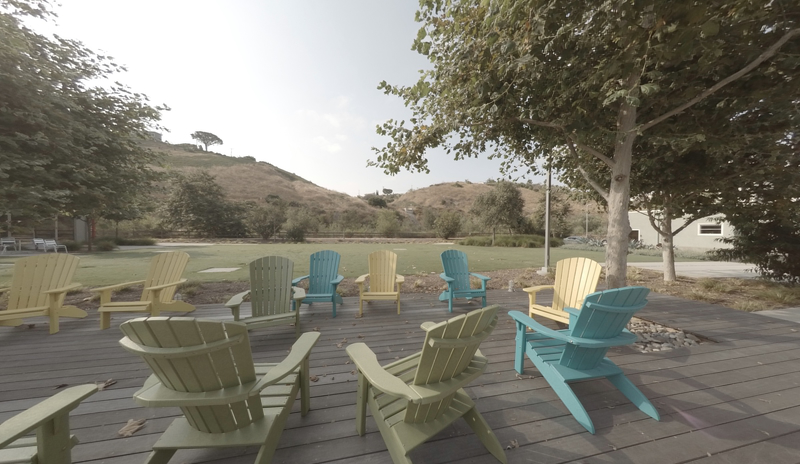} &
\includegraphics[width=0.3\linewidth,trim={300 46 300 234},clip=true]{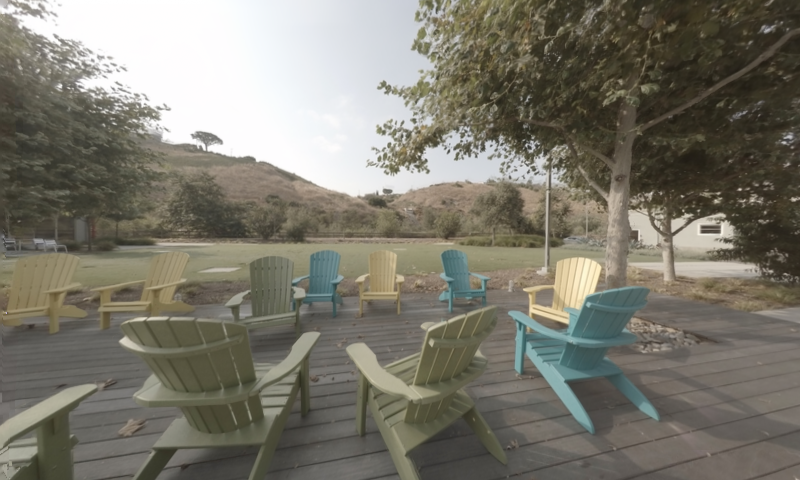} &
\includegraphics[width=0.3\linewidth,trim={300 46 300 234},clip=true]{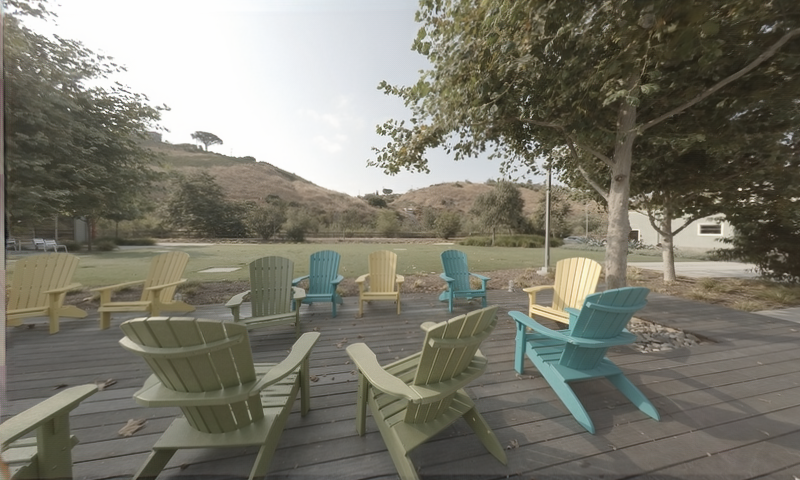} 
\\
\small GT & \small Soft3D \cite{penner2017soft} & \small DeepView \cite{flynn2019deepview}
\end{tabular}
\begin{tabular}{@{}c@{\hskip0.4em}c@{}}
\includegraphics[width=0.3\linewidth,trim={300 30 300 234},clip=true]{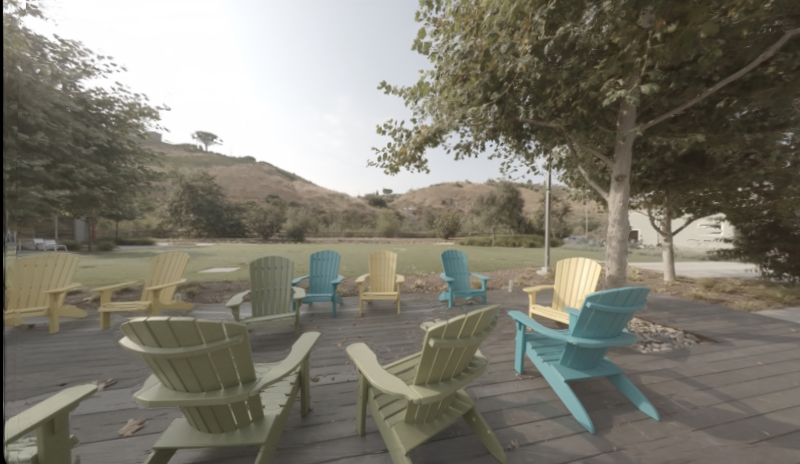} &
\includegraphics[width=0.3\linewidth,trim={300 30 300 234},clip=true]{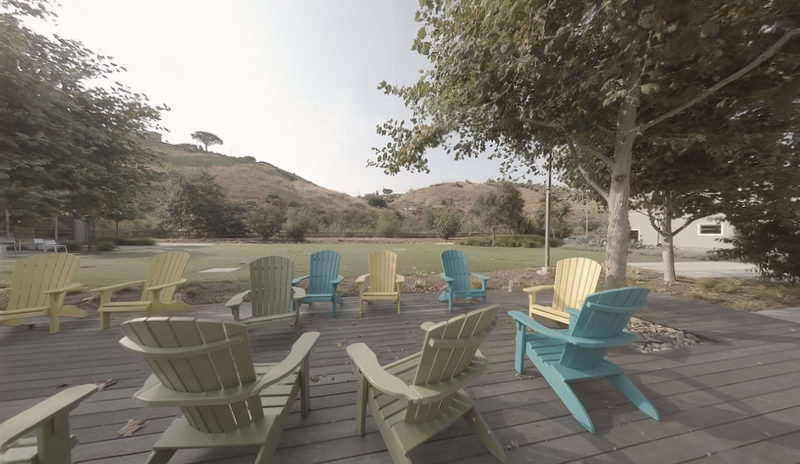} 
\\
\small RGSA \cite{volker2020learning} & \small Proposed
\end{tabular}
}
\caption{\color{black} Crop of a synthesized central view of one scene of the Spaces test set, compared with the ground truth (GT) and state-of-the-art methods.}
\label{fig:comparison_spaces}
\end{figure}

\textbf{Comparison against the state of the art.} 
Table \ref{tab:sota} compares our method with Soft3D \cite{penner2017soft}, DeepView \cite{flynn2019deepview} and RGSA~\cite{volker2020learning}. Learning methods have been trained with the training data associated to the considered evaluation set. For RGSA, we used the authors code and trained the network using the same procedure as with our method. The codes of Soft3D and DeepView are not publicly available and we computed the metrics with the provided results on the Spaces dataset. At inference, DeepView uses 80 planes, instead of 60 for RGSA and our method (due to memory limitations). DeepView does not share parameters between refinement iterations and uses a larger model that processes per-view features at different stages of the network. While their approach provides accurate MPIs, it requires a lot of memory and is not practical to train. 
To lower complexity, RGSA and our method share the weights across iterations and reduce the features along the view dimension. 
Still, contrary to RGSA, our network includes the deep features $F_n$ as input and estimates all MPI channels.
Our solution is slightly less accurate than DeepView, but better than Soft3D and RGSA.
Fig.~\ref{fig:comparison_id} shows the improvements of our model over RGSA on two examples of the ID test set. Apart from superior geometry predictions, our results are noticeably sharper. Finally, Fig. \ref{fig:comparison_spaces} shows that even using a lower number of planes our approach is visually comparable to DeepView, with a sharp synthesized view, while Soft3D and RGSA produce blurred results.

\section{Conclusion} \label{sec:conclusions}
We have proposed a new method to produce compact and adaptive MPIs. Our strategy allows to render new views with high accuracy and limited memory footprint. Adapting the depths to the scene during inference we optimize the available memory and constraining the MPI to be compact we force the network to only keep the important information that should be compressed. We believe that our compact and adaptive MPIs are making a big step forward towards the deployment of volumetric technologies in the immersive realm. 


\bibliographystyle{IEEEbib}
\bibliography{refs}

\begin{thebibliography}{10}

\bibitem{zhou2018stereo}
T.~Zhou, R.~Tucker, J.~Flynn, G.~Fyffe, and N.~Snavely,
\newblock ``Stereo magnification: Learning view synthesis using multiplane
  images,''
\newblock {\em ACM Transactions on Graphics}, vol. 37, no. 4, pp. 1--12, 2018.

\bibitem{srinivasan2019pushing}
P.~P. Srinivasan, R.~Tucker, J.~T. Barron, R.~Ramamoorthi, R.~Ng, and
  N.~Snavely,
\newblock ``Pushing the boundaries of view extrapolation with multiplane
  images,''
\newblock in {\em Proceedings of the IEEE Conference on Computer Vision and
  Pattern Recognition}, 2019, pp. 175--184.

\bibitem{mildenhall2019local}
B.~Mildenhall, P.~P. Srinivasan, R.~Ortiz-Cayon, N.~K. Kalantari,
  R.~Ramamoorthi, R.~Ng, and A.~Kar,
\newblock ``Local light field fusion: Practical view synthesis with
  prescriptive sampling guidelines,''
\newblock {\em ACM Transactions on Graphics}, vol. 38, no. 4, pp. 1--14, 2019.

\bibitem{flynn2019deepview}
J.~Flynn, M.~Broxton, P.~Debevec, M.~DuVall, G.~Fyffe, R.~Overbeck, N.~Snavely,
  and R.~Tucker,
\newblock ``Deepview: View synthesis with learned gradient descent,''
\newblock in {\em Proceedings of the IEEE Conference on Computer Vision and
  Pattern Recognition}, 2019, pp. 2367--2376.

\bibitem{broxton2020immersive}
M.~Broxton, J.~Flynn, R.~Overbeck, D.~Erickson, P.~Hedman, M.~DuVall,
  J.~Dourgarian, J.~Busch, M.~Whalen, and P.~Debevec,
\newblock ``Immersive light field video with a layered mesh representation,''
\newblock {\em ACM Transactions on Graphics}, vol. 39, no. 4, pp. 86:1--86:15,
  2020.

\bibitem{volker2020learning}
T.~V\"olker, G.~Boisson, and B.~Chupeau,
\newblock ``Learning light field synthesis with multi-plane images: scene
  encoding as a recurrent segmentation task,''
\newblock in {\em Proceedings of the IEEE International Conference on Image
  Processing}, 2020.

\bibitem{ba2016layer}
J.~L. Ba, J.~R. Kiros, and G.~E. Hinton,
\newblock ``Layer normalization,''
\newblock {\em arXiv preprint arXiv:1607.06450}, 2016.

\bibitem{ronneberger2015u}
O.~Ronneberger, P.~Fischer, and T.~Brox,
\newblock ``U-{N}et: Convolutional networks for biomedical image
  segmentation,''
\newblock in {\em Proceedings of the International Conference on Medical Image
  Computing and Computer-Assisted Intervention}. Springer, 2015, pp. 234--241.

\bibitem{kingma2015adam}
D.~P. Kingma and J.~Ba,
\newblock ``Adam: A method for stochastic optimization,''
\newblock in {\em Proceedings of the International Conference on Learning
  Representations}, 2015.

\bibitem{glorot2010understanding}
X.~Glorot and Y.~Bengio,
\newblock ``Understanding the difficulty of training deep feedforward neural
  networks,''
\newblock in {\em Proceedings of the International Conference on Artificial
  Intelligence and Statistics}, 2010, pp. 249--256.

\bibitem{wang2004image}
Z.~Wang, A.~C. Bovik, H.~R. Sheikh, and E.~P. Simoncelli,
\newblock ``Image quality assessment: from error visibility to structural
  similarity,''
\newblock {\em IEEE Transactions on Image Processing}, vol. 13, no. 4, pp.
  600--612, 2004.

\bibitem{zhang2018unreasonable}
R.~Zhang, P.~Isola, A.~A. Efros, E.~Shechtman, and O.~Wang,
\newblock ``The unreasonable effectiveness of deep features as a perceptual
  metric,''
\newblock in {\em Proceedings of the IEEE Conference on Computer Vision and
  Pattern Recognition}, 2018, pp. 586--595.

\bibitem{sabater2017dataset}
N.~Sabater, G.~Boisson, B.~Vandame, P.~Kerbiriou, F.~Babon, M.~Hog, R.~Gendrot,
  T.~Langlois, O.~Bureller, A.~Schubert, and V.~Alli\'e,
\newblock ``Dataset and pipeline for multi-view light-field video,''
\newblock in {\em Proceedings of the IEEE Conference on Computer Vision and
  Pattern Recognition Workshops}, 2017, pp. 30--40.

\bibitem{markley2007averaging}
F.~L. Markley, Y.~Cheng, J.~L. Crassidis, and Y.~Oshman,
\newblock ``Averaging quaternions,''
\newblock {\em Journal of Guidance, Control, and Dynamics}, vol. 30, no. 4, pp.
  1193--1197, 2007.

\bibitem{penner2017soft}
E.~Penner and L.~Zhang,
\newblock ``Soft 3{D} reconstruction for view synthesis,''
\newblock {\em ACM Transactions on Graphics}, vol. 36, no. 6, pp. 1--11, 2017.

\end{thebibliography}

\end{document}